\documentclass[conference]{IEEEtran}
\IEEEoverridecommandlockouts

\usepackage{cite}
\usepackage{amsmath,amssymb,amsfonts}
\usepackage{algorithmic}
\usepackage{graphicx}
\usepackage{textcomp}
\usepackage{xcolor}
\usepackage{hyperref}
\usepackage{booktabs} 
\usepackage{url}

\usepackage{bigstrut,multirow,rotating,booktabs}
\usepackage{array}
\usepackage{tabularx}
\usepackage{textcomp}
\usepackage{subcaption}
\usepackage{caption}
\usepackage{times}
\usepackage{latexsym}
\usepackage{amssymb}
\usepackage{bbding}
\usepackage{inconsolata}
\hypersetup{hidelinks=true}
\usepackage{makecell}
\usepackage{float}
\usepackage{colortbl}
\usepackage{multirow}
\usepackage{xcolor}

\def\BibTeX{{\rm B\kern-.05em{\sc i\kern-.025em b}\kern-.08em
    T\kern-.1667em\lower.7ex\hbox{E}\kern-.125emX}}
\begin{document}

\title{The Multi-Round Diagnostic RAG Framework for Emulating Clinical Reasoning}

\author{
\IEEEauthorblockN{
        Penglei Sun,
        Yixiang Chen,
        Xiang Li, and
        Xiaowen Chu\IEEEauthorrefmark{1}\thanks{*Corresponding author.}
    }
    \IEEEauthorblockA{
        The Hong Kong University of Science and Technology (Guangzhou), Guangzhou, China\\
        \{psun012, ychen416, xli906, xwchu\}@connect.hkust-gz.edu.cn 
    }

}

\maketitle

\begin{abstract}
In recent years, accurately and quickly deploying medical large language models (LLMs) has become a trend. Among these, retrieval-augmented generation (RAG) has garnered attention due to rapid deployment and privacy protection. 
However, the challenge hinder the practical deployment of RAG for medical diagnosis: the semantic gap between colloquial patient descriptions and the professional terminology within medical knowledge bases. 
We try to address the challenge from the \underline{data perspective} and the \underline{method perspective}.
First, to address the semantic gap in existing knowledge bases, we construct \underline{DiagnosGraph}, a generalist knowledge graph covering both modern medicine and Traditional Chinese Medicine. 
It contains $876$ common diseases with the graph of $7,997$ nodes and $37,201$ triples. 
To bridge the gap between colloquial patient narratives and academic medical knowledge, DiagnosGraph also introduces $1,908$ medical record by formalizing the patient chief complaint and proposing a medical diagnosis.
Second, we introduce the \underline{M}ulti-\underline{R}ound \underline{D}iagnostic \underline{RAG} (\underline{MRD-RAG}) framework. 
It utilizes a multi-round dialogue to refine diagnostic possibilities, emulating the clinical reasoning of a physician.
Experiments conducted on four medical benchmarks, with evaluations by human physicians, demonstrate that MRD-RAG enhances the diagnostic performance of LLMs, highlighting its potential to make automated diagnosis more accurate and human-aligned.
The codebase and supplementary materials can be found at our page\footnote{\textit{\url{https://sites.google.com/view/mrd-rag}}}.
\end{abstract}


\section{Introduction}
In recent years, medical large language models (LLMs) have emerged as crucial tools in addressing the growing demands on healthcare systems \cite{liu2024survey}. 
Two primary approaches exist for integrating medical knowledge into LLMs: parameter modification through finetuning and prompt-based integration via Retrieval-Augmented Generation (RAG) \cite{balaguer2024rag,ovadia2023fine}.
While finetuning is effective, it is often constrained by substantial computational demands \cite{yang2024zhongjing,bao2023disc}. 
In contrast, RAG has garnered attention due to the rapid deployment, privacy protection, and the ability to incorporate the latest knowledge \cite{fan2024survey,zhu2022multi}. 
This is achieved by first retrieving relevant information from a knowledge base and then augmenting the model's prompt to generate more accurate responses. 
The capacity for real-time knowledge updates makes RAG particularly well-suited for a knowledge-intensive domain like healthcare.

\begin{figure}[htbp!]
    \centering
    \includegraphics[width=1.05\linewidth]{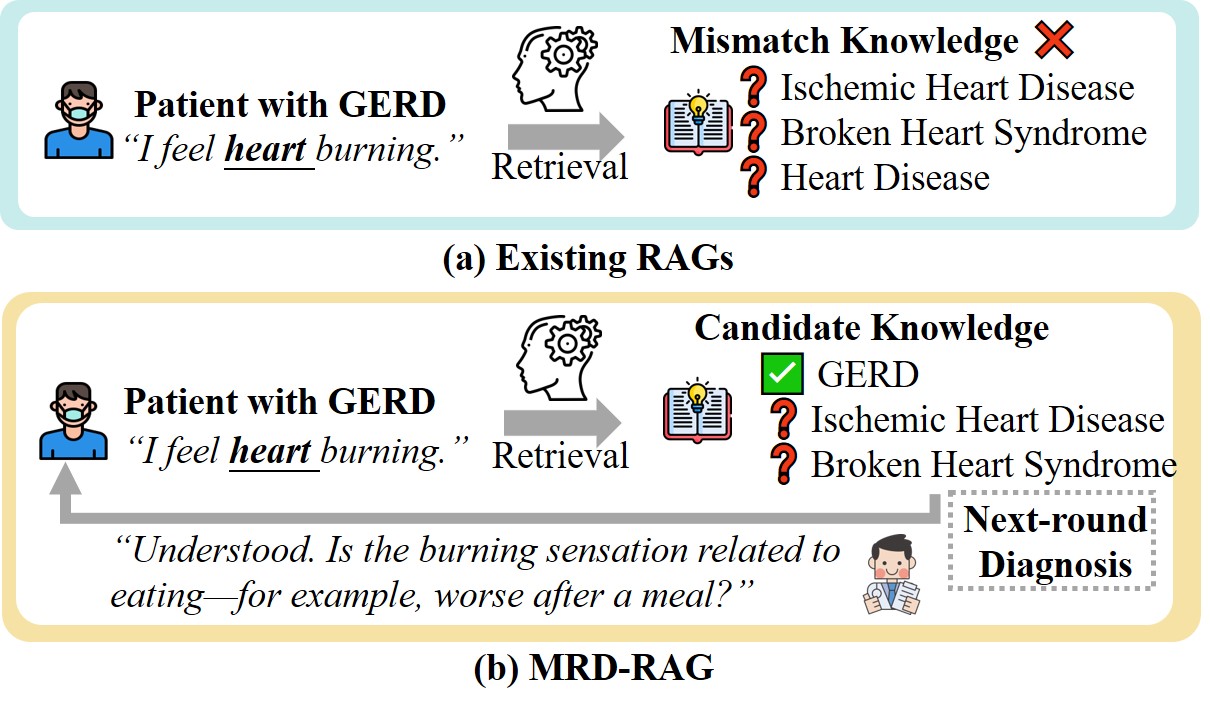}
    \caption{Comparison between MRD-RAG with existing RAGs. 
    GERD denotes Gastroesophageal Reflux Disease, a common stomach condition with heart symptom.}
    \label{fig:Overall}
\end{figure}

Despite its potential, the challenges compromise the practical application of RAG in medical diagnosis:\textbf{the semantic gap} between the colloquial language of patients and the formal terminology of medical knowledge bases.
As shown in Fig~\ref{fig:Overall} (a), a patient might describe their symptom using a colloquial phrase like "\textit{I feel heart burning}" or "\textit{a burning feeling in my chest.}" 
A standard RAG system, relying on semantic similarity, can be misled by patient descriptions. 
For instance, a query containing "\textit{heart burning}" might cause the system to retrieve mismatch knowledge about heart conditions (e.g., "\textit{Ischemic Heart Disease}"), while neglecting the correct but less obvious stomach diagnosis (e.g., "\textit{Gastroesophageal Reflux Disease}").  
In clinical practice, human physicians overcome this challenge by employing a multi-round, reasoning process. 
Faced with a complaint like "\textit{heart burning}," they don't perform a simple keyword search. 
Instead, they initiate a dialogue, asking targeted questions about the pain's characteristics with medical reports and other symptoms to distinguish between numerous possibilities through multi-round diagnosis.
This ability to dynamically probe and refine a diagnosis contrasts with existing RAG models, which typically lack the capability for such iterative, context-aware inquiry~\cite{li2023chatdoctor,xiong2024benchmarking,wang2024jmlr,jiang2023think}.

To achieve a doctor-like and multi-round diagnostic RAG framework, we explore the knowledge-enhanced LLM from the data and the method perspective, as shown Fig~\ref{fig:Overall} (b).
\underline{From the data perspective}, current medical knowledge bases lack both detailed diagnostic information and the semantic mechanisms to connect colloquial patient expressions with professional terminology~\cite{yang2021mining,li2020kghc,li2023meddm,zhao2023construction,duan2025research,zhang2023etcm}.
We introduce \textbf{DiagnosGraph}, a knowledge graph designed to solve these issues with two key features. First, it serves as a knowledge base, integrating detailed diagnostic information for $\textbf{876}$ common diseases from multiple sources into a graph of $\textbf{7,997}$ nodes and $\textbf{37,201}$ triples. 
Second, it bridges the semantic gap by formalizing patient narratives into a dataset of $\textbf{1,908}$ medical records, each mapping a chief complaint to a potential diagnosis.

\underline{From the method perspective}, we introduce the \textbf{M}ulti-Round \textbf{D}iagnostic \textbf{RAG} (\textbf{MRD-RAG}) framework, which emulates a physician's clinical reasoning through three core modules: a Retriever, an Analyzer, and a Doctor.
This process begins when a patient presents a complaint, such as a "\textit{burning feeling}."
First, the Retriever identifies multiple candidate diseases with knowledge from DiagnosGraph (e.g., migraine, tension headache, sinusitis).
Next, mimicking a physician's differential diagnosis, the Analyzer evaluates the relationships between these candidates and assesses the patient's symptoms against each one.
Finally, the Doctor module synthesizes this analysis to determine the next best action: either posing a targeted follow-up question to distinguish between top hypotheses (e.g., asking about "\textit{nasal congestion}" to probe for sinusitis) or delivering a final diagnosis.

To evaluate the effectiveness of MRD-RAG, we conduct experiments with existing LLMs and RAGs on four public benchmarks using GPT scores and human doctors. 
The GPT scores indicate that MRD-RAG improves the diagnostic performance of LLMs without RAG and with exsiting RAG by an average of $\textbf{9.4\%}$ and $\textbf{6.0\%}$, respectively. 
We also ask human doctors to evaluate the diagnostic performance of our framework against LLMs without RAG and using RAG respectively. 
Our framework achieves the improvement of $\textbf{21.75\%}$ and $\textbf{18\%}$, respectively, over these two methods.

Our main contributions can be summarized as follows:

\begin{enumerate}
    \item We introduce DiagnosGraph, a novel medical knowledge graph. It not only provides detailed diagnostic information for $\textbf{876}$ common diseases with $\textbf{7,997}$ nodes and $\textbf{37,201}$ triples but also bridges the semantic gap by mapping colloquial patient complaints to formal medical diagnoses through medical record.

    \item We propose MRD-RAG, a conversational diagnostic framework that emulates a physician's reasoning. By iteratively performing differential diagnosis and targeted inquiry, the framework dynamically refines diagnostic possibilities, moving beyond the static, single-round limitations of existing RAG models.

    \item We conduct extensive experiments and human doctor evaluations that demonstrate the effectiveness and superiority of our approach. Results show that MRD-RAG enhances the diagnostic performance of various LLMs over $\textbf{6.0 \%}$, achieving state-of-the-art results that are highly aligned with clinical expertise.
\end{enumerate}

\section{Related Work}

\subsection{Retrieval-Augmented Generation}

Recent advancements in Retrieval-Augmented Generation (RAG) have explored complex dialogues.
Some methods aim to enhance retrieval within conversations, for instance by reformulating queries for better semantic alignment (HyKGE~\cite{jiang2023think}) or by handling dialogue context (RagPULSE~\cite{huang2024tool}, ConvRAG~\cite{ye2024boosting}). 
Others employ multi-retrieval strategies for complex reasoning, such as multi-hop question answering (ToG~\cite{sun2023think}) or iterative query refinement (IM-RAG~\cite{yang2024rag}). 
However, these approaches fall short of emulating a physician's diagnostic reasoning. 
They typically focus on fact-based retrieval or surface-level dialogue coherence, rather than performing differential diagnosis among candidate diseases and engaging in strategic, multi-round inquiry.

\subsection{Medical LLM}

Medical LLMs focus on enhancing models' capabilities across various medical tasks, such as medical commonsense Q\&A, medication consultation, and clinical diagnosis. 
DISC-MedLLM \cite{bao2023disc}, DoctorGLM \cite{xiong2023doctorglm} construct high-quality medical multi-round dialogue datasets and perform supervised fine-tuning on LLMs. 
HuatuoGPT-II \cite{chen2023huatuogptii} and Baichuan-M1~\cite{wang2025baichuan} combines continued pre-training and supervised fine-tuning stages into a single process so as to solve the problem of inconsistent data distribution in medical LLM training.  
Furthermore, some studies apply RAG to general LLMs to enhance reliability. 
Self-BioRAG \cite{jeong2024improving} introduces Self-RAG \cite{asai2023self} into the biomedical field to dynamically decide whether retrieval is needed. 
MedDM \cite{li2023meddm} proposes ``clinical guidance tree'' (CGT) and let the LLM reason on it. 
Our framework analyzes the interconnections and differences of retrieved potential diseases that patients may suffer, enhancing the performance of LLMs in proactive inquiry and diagnosis.

\section{Method} \label{Method}

\begin{figure*}[t]
    \centering
    \includegraphics[width=1.0\linewidth]{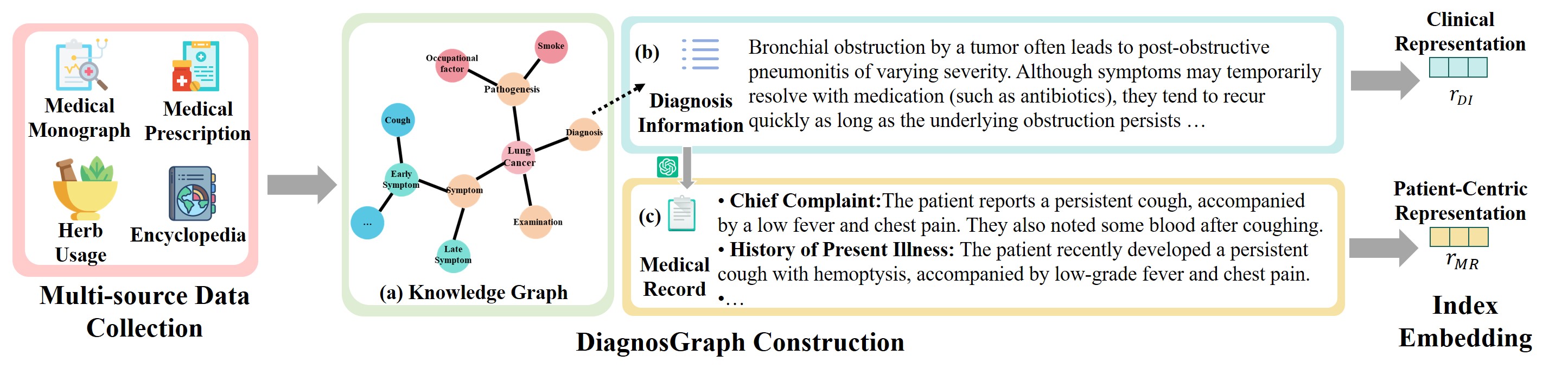}
    \caption{DiagnosGraph construction process.}
    \label{fig:Database}
\end{figure*}


The MRD-RAG pipeline is modeled as an interactive dialogue between our proposed doctor agent and a patient. 
In each conversational round $i$, the doctor agent processes the patient's latest complaint, $x_{patient}^i$, within the context of the dialogue history, $H^{i-1}=[x_{patient}^1, q_{doctor}^1, ..., x_{patient}^{i-1}]$.
First, the agent retrieves a relevant knowledge subset $K^i$ from the knowledge base. 
Based on the history $H^{i-1}$ and the retrieved knowledge $K^i$, the agent then decides its next action: either posing a targeted question, $q_{doctor}^i$, to elicit more specific information, or concluding with a diagnosis.
The primary challenge in this process is the semantic gap between the patient's colloquial descriptions ($x_{patient}^i$) and the formal terminology within the expert knowledge base. 
To address this, we introduce DiagnosGraph, a knowledge graph specifically designed to bridge this gap, as shown in Fig~\ref{fig:Database}.

\subsection{DiagnosGraph}
\subsubsection{Data Collection}

To construct DiagnosGraph, we collect multi-source data~\footnote{https://www.yixue.com/, https://www.dayi.org.cn/}.  
The DiagnosGraph contains three parts:

\noindent$\bullet$~\textbf{Knowledge Graph}. First, we build a structured knowledge graph to serve as the data foundation. 
Following the World Health Organization's International Classification of Diseases (ICD) schema, this graph covers $876$ common diseases from both Modern Medicine and Traditional Chinese Medicine (TCM). 
Each disease connects to various attributes, such as Symptom, Examination, and Pathogenesis. 
This comprises $7,997$ nodes and $37,201$ triples.

\noindent$\bullet$~\textbf{Diagnosis Information}. 
Building upon this structure, we associate each disease with clinical diagnostic information. 
For instance, for the ``\textit{Lung Cancer}", we include its detailed clinical presentations with diagnostic criteria, and treatment variations. 

\noindent$\bullet$~\textbf{Medical Records}.
We make this professional knowledge applicable to real-world patient dialogue through medical records.
Addressing the privacy limitations of real medical data, we use the structured and professional information from the first two components as a seed for GPT-4o to generate simulated clinical records supervised by medical experts.
The records are both clinically accurate and authentically patient-like.

As shown in Table~\ref{tab:EKB}, we compare our knowledge base with existing medical knowledge bases. 
Compared to existing ones, which lack detailed diagnostic information and medical records, DiagnosGraph is the first to cover both modern medicine and TCM diseases in a generalist domain with fine-grained diagnostic details.

\begin{table*}[t]
  \centering
  \caption{Comparison of medical knowledge bases.}
  \scalebox{1.0}{
\begin{tabular}{@{}c|cccccc@{}}
\toprule
Name                                                          & \begin{tabular}[c]{@{}c@{}}TCM\\ Disease\end{tabular} & \begin{tabular}[c]{@{}c@{}}Modern Medicine\\ Disease\end{tabular} & Medical Record              & \begin{tabular}[c]{@{}c@{}}Diagnosis\\ Information\end{tabular} & Domain     & Scale                                                                                          \\ \midrule
CMeKG~\cite{byambasuren2019preliminary} & \XSolidBrush                           & \Checkmark                                         & \XSolidBrush & \XSolidBrush                                     & Generalist & over 100 millions triples                                                                      \\
DKD-TCMKG~\cite{zhao2023construction}   & \Checkmark                             & \XSolidBrush                                       & \XSolidBrush & \XSolidBrush                                     & Specialist & 903 nodes                                                                                      \\
WZQ-TCMKG~\cite{duan2025research}       & \Checkmark                             & \Checkmark                                         & \Checkmark   & \XSolidBrush                                     & Generalist & 679 medical records                                                                            \\
RD-MKG~\cite{li2020real}                & \XSolidBrush                           & \Checkmark                                         & \XSolidBrush & \XSolidBrush                                     & Generalist & 22,508 entities,579,094 triples                                                                \\
StrokeKG~\cite{yang2021mining}          & \XSolidBrush                           & \Checkmark                                         & \XSolidBrush & \XSolidBrush                                     & Specialist & 46k nodes, 157k triples                                                                        \\
KGHC~\cite{li2020kghc}                  & \XSolidBrush                           & \Checkmark                                         & \XSolidBrush & \XSolidBrush                                     & Specialist & 5028 nodes, 13,296 triples                                                                     \\
ETCM2.0~\cite{zhang2023etcm}            & \Checkmark                             & \Checkmark                                         & \XSolidBrush & \XSolidBrush                                     & Generalist & over 100k entities                                                                             \\
CGT~\cite{li2023meddm}                  & \XSolidBrush                           & \Checkmark                                         & \XSolidBrush & \XSolidBrush                                     & Generalist & 1,202 nodes                                                                                    \\ \midrule
DiagnosGraph (ours)                                           & \Checkmark                             & \Checkmark                                         & \Checkmark   & \Checkmark                                       & Generalist & \begin{tabular}[c]{@{}c@{}}7,997 nodes and 37,201 triples,\\ 1908 medical records\end{tabular} \\ \bottomrule
\end{tabular}
    }
  \label{tab:EKB}%
\end{table*}%

\subsubsection{Indexing} \label{Indexing}
To bridge the semantic gap between a patient's colloquial utterances and formal diagnostic texts ($t_{DI}^{i}$), we employ a dual-representation strategy for each disease $d^{i}$. (1) Clinical Representation ($r_{DI}^{i}$): We first compute a representation directly from the formal diagnostic text: $r_{DI}^i=Embedding(t_{DI}^i)$. 
(2) Patient-Centric Representation ($r_{MR}^i$): To capture the patient's perspective, we then embed  patient narrative ($t_{MR}^i$) to create a second vector: $r_{MR}^i=Embedding(t_{MR}^i)$.

\subsection{Multi-round Diagnostic RAG Framework}

\begin{figure}
    \centering
    \includegraphics[width=1\linewidth]{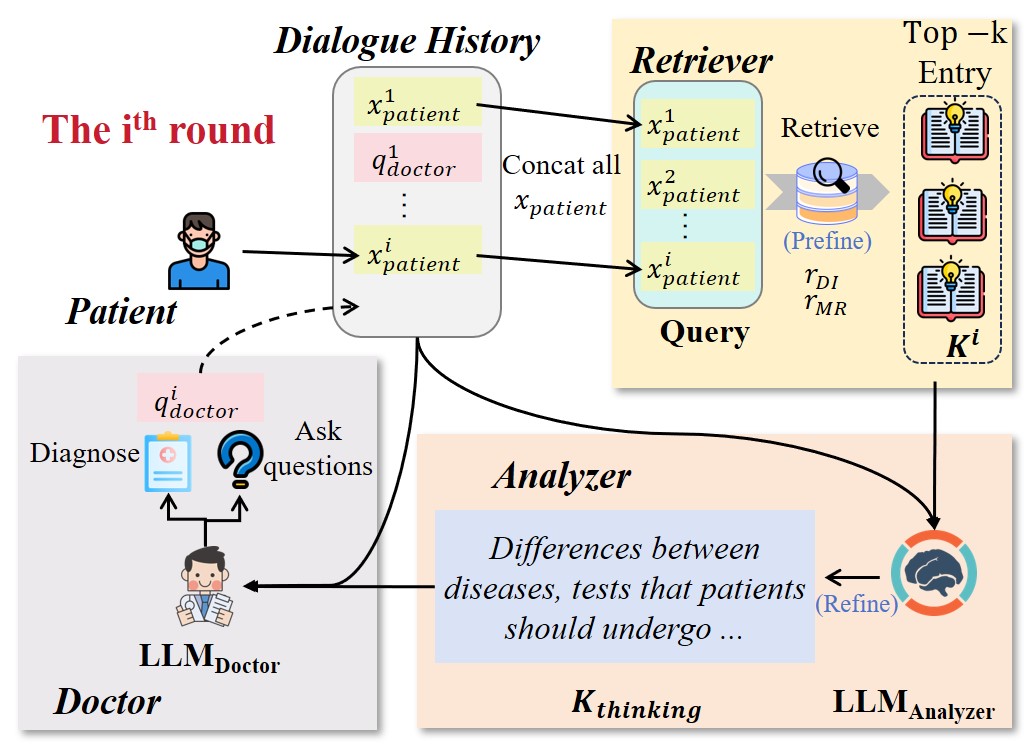}
    \caption{The pipeline of MRD-RAG.}
    \label{fig:Pipeline}
\end{figure}

As shown in Figure \ref{fig:Pipeline}, MRD-RAG comprises three main modules: \textbf{Retriever}, \textbf{Analyzer}, and \textbf{Doctor}.

\noindent$\bullet$~\textbf{Retriever Module.}
The Retriever's goal is to identify candidate diseases based on the patient's own words, minimizing bias from preliminary machine inferences.
The query is constructed by concatenating the patient's entire dialogue history: $H_{patient}=[x_{patient}^1, … ,x_{patient}^i]$.
The query embedding, $r_d=Embedding(H_{patient})$, is used to search across both $Index_{DI}$ and $Index_{MR}$ through cosine similarity.
To improve efficiency, a gating LLM first assesses if the latest patient utterance adds new diagnostic information. 
Retrieval is skipped if the turn is non-informative (e.g., "I see"), preventing redundant processing.
The module outputs the the top-k retrieved diseases, forming the raw knowledge base $K$ for the next stage.

\noindent$\bullet$~\textbf{Analyzer Module.}
The Analyzer acts as a clinical reasoning engine through synthesizes the raw knowledge ($K^i$) and dialogue history ($H^{i} = [x_{patient}^1, q_{doctor}^1, …,x_{patient}^i]$).
It first synthesizes the dense information in $K^i$ through comparing the candidate diseases to identify key distinguishing features and crucial overlaps.
Then it formulates a clear diagnostic strategy and outputs the summary knowledge $K_{thinking}$, such as which specific symptoms to ask about next to resolve ambiguity, i.e., $K_{thinking} = LLM_{Analyzer}(H^{i}, K^i)$.

\noindent$\bullet$~\textbf{Doctor Module.}
Using the $K_{thinking}$ as its guide, the Doctor module formulates the final reply $x_{doctor}^n = LLM_{Doctor}(H^{i}, K_{thinking})$. 
Based on the strategy provided by the Analyzer, it will execute one of two actions:
(1) Diagnose: If $K_{thinking}$indicates that sufficient evidence has been gathered, it will deliver a clear diagnosis.
(2) Inquire: If ambiguities remain, it will pose another question based on the $K_{thinking}$ in the next round dialogue.

\section{Experiments}

\subsection{Experimental Setup}

\subsubsection{Evaluation Dataset}\label{Evaluation Dataset}

We utilize four datasets for evaluation, including two publicly available datasets and two constructed by us. 
CMB-Clin \cite{wang2023cmb} is one of the publicly available subsets of a Chinese medical benchmark, containing $74$ complex modern medical clinical diagnosis cases. 
TCM-SD \cite{mucheng2022tcm} is the first public benchmark for TCM syndrome differentiation, and it normalizes all TCM syndrome names. 
From its test set, we randomly select $100$ cases (TCM-SD-100) for evaluation. 
Besides, we collect $609$ modern medicine patient cases and $130$ TCM patient cases from doctor clinic records as the MM-cases and TCM-cases respectively.

\subsubsection{Implementation Details}  
During the multi-round dialogue, we consistently employ GPT-4o-mini to play the role of the patient and engage in dialogue with different methods. 
As described in Section \ref{Method}, in each round of dialogue, the patient converses based on their medical information $I_{patient}$ and the dialogue history. To simulate the real-world patient consultation scenario, we encourage the patient to avoid directly revealing their real disease name and to refrain from disclosing excessive information in a single utterance. 
As for the embedding model, we uniformly use text-embedding-3-small \footnote{https://platform.openai.com/docs/guides/embeddings} in the indexing and retrieval stages, which is a text embedding model for a general domain. 
We retrieve top five relevant knowledge from the DiagnosGraph ($top\text{-}k=5$).


\subsubsection{Baseline}
We select the following four groups of methods as baselines: 

\noindent \textbf{$\bullet$ w/o RAG:} This group includes two open-source general LLMs: Qwen2-7B-Instruct and Qwen2-72B-Instruct-AWQ~\cite{qwen2}, as well as two popular closed-source LLMs: GPT-4o-mini and GPT-4o.
Additionally, we choose medical LLM including DISC-MedLLM \cite{bao2023disc}, HuatuoGPT-II-7B \cite{chen2023huatuogptii} and Baichuan-M1-14B-Instruct \cite{wang2025baichuan} for comparison. Both of them have been trained on different specialized medical datasets.

\noindent \textbf{$\bullet$ Standard RAG:} To validate the effectiveness of MRD-RAG, we also compare it with Langchain RAG. 
Specifically, the LLM only utilize the retrieved knowledge. 
For fair comparison, we also select Qwen2-7B-Instruct, Qwen2-72B-Instruct-AWQ, GPT-4o-mini and Baichuan-M1-14B-Instruct as base models.

\noindent \textbf{$\bullet$ Medical RAG:} We compare the medicine domain RAG method RagPULSE \cite{huang2024tool} which utilizes LLMs to summarize keywords in dialogue to form queries for retrieval.

\begin{table*}[htbp]
  \centering
  \caption{GPT-4o evaluation for each method and model (1-5 points). *The results are from our code replication.}
    \resizebox{0.9\linewidth}{!}{
\begin{tabular}{@{}c|c|cccc|c@{}}
\toprule
Method                                                                         & Model                  & CMB-Clin & MM-Cases & TCM-SD-100 & TCM-Cases & Average                                                      \\ \midrule
\multirow{4}{*}{\begin{tabular}[c]{@{}c@{}}General LLM\\ w/o RAG\end{tabular}} & Qwen2-7B-Instruct      & 3.12     & 2.85     & 2.62       & 3.32      & \cellcolor[rgb]{ .984, .757, .769} 2.98 \\
                                                                               & Qwen2-72B-Instruct-AWQ & 3.24     & 2.93     & 2.82       & 3.69      & \cellcolor[rgb]{ .98, .675, .682} 3.17  \\
                                                                               & GPT-4o-mini            & 2.91     & 2.71     & 2.97       & 3.24      & \cellcolor[rgb]{ .984, .765, .776} 2.96 \\
                                                                               & GPT-4o                 & 2.99     & 2.85     & 2.99       & 3.42      & \cellcolor[rgb]{ .984, .722, .733} 3.06 \\ \midrule
\multirow{3}{*}{\begin{tabular}[c]{@{}c@{}}Medical LLM\\ w/o RAG\end{tabular}} & DISC-MedLLM            & 2.74     & 2.49     & 2.22       & 2.33      & \cellcolor[rgb]{ .988, .988, 1} 2.45    \\
                                                                               & HuatuoGPT-II-7B        & 2.82     & 2.69     & 2.35       & 2.86      & \cellcolor[rgb]{ .988, .89, .902} 2.68  \\
                                                                               & Baichuan-M1-14B        & 2.85     & 2.73     & 2.75       & 3.13      & \cellcolor[rgb]{ .984, .808, .82} 2.86  \\ \midrule
\multirow{4}{*}{Standard RAG}                                                  & Qwen2-7B-Instruct      & 3.15     & 2.96     & 3.03       & 3.90      & \cellcolor[rgb]{ .98, .635, .643} 3.26  \\
                                                                               & Qwen2-72B-Instruct-AWQ & 3.21     & 3.27     & 3.16       & 3.95      & \cellcolor[rgb]{ .976, .518, .529} 3.40 \\
                                                                               & GPT-4o-mini            & 3.12     & 2.92     & 3.20       & 3.81      & \cellcolor[rgb]{ .98, .635, .643} 3.26  \\
                                                                               & Baichuan-M1-14B        & 2.70     & 2.68     & 2.49       & 3.59      & \cellcolor[rgb]{ .984, .804, .816} 2.87 \\ \midrule
Medical RAG                                                                    & RagPULSE-20B$^{*}$     & 3.22     & 3.03     & 3.12       & 3.55      & \cellcolor[rgb]{ .98, .647, .655} 3.23  \\ \midrule
\multirow{4}{*}{MRD-RAG-DI}                                                    & Qwen2-7B-Instruct      & 3.23     & 3.23     & 3.14       & 3.98      & \cellcolor[rgb]{ .98, .576, .584} 3.39  \\
                                                                               & Qwen2-72B-Instruct-AWQ & 3.26     & 3.35     & 3.19       & 3.99      & \cellcolor[rgb]{ .976, .549, .561} 3.45 \\
                                                                               & GPT-4o-mini            & 3.14     & 3.24     & 3.19       & 4.12      & \cellcolor[rgb]{ .98, .565, .573} 3.42  \\
                                                                               & Baichuan-M1-14B        & 3.18     & 3.22     & 2.99       & 3.75      & \cellcolor[rgb]{ .98, .624, .635} 3.28  \\ \midrule
\multirow{4}{*}{MRD-RAG-MR}                                                    & Qwen2-7B-Instruct      & 3.59     & 3.83     & 3.26       & 4.36      & \cellcolor[rgb]{ .973, .412, .42} 3.76  \\
                                                                               & Qwen2-72B-Instruct-AWQ & 3.62     & 3.77     & 3.09       & 4.43      & \cellcolor[rgb]{ .976, .427, .435} 3.73 \\
                                                                               & GPT-4o-mini            & 3.53     & 3.75     & 3.23       & 4.42      & \cellcolor[rgb]{ .976, .427, .435} 3.73 \\
                                                                               & Baichuan-M1-14B        & 3.26     & 3.82     & 2.82       & 4.02      & \cellcolor[rgb]{ .976, .537, .545} 3.48 \\ \bottomrule
\end{tabular}
}%
  \label{tab:gpt4 scores}%
\end{table*}%

\subsubsection{Metric}\label{Metric}

All methods dialogue with the patient. Subsequently, we establish different evaluation metrics to compare different methods in the dialogues. 

\noindent \textbf{$\bullet$ GPT Evaluation Metric:} Some studies indicate that powerful LLMs can achieve high alignment with human doctors' judgments \cite{bao2023disc,yang2024zhongjing,zhang2023huatuogpt}. 
We employ GPT-4o to simulate human doctors for evaluation as existing works\cite{bao2023disc,yang2024zhongjing,zhang2023huatuogpt}. 
Specifically, we provide the patient information and the corresponding dialogues of different methods in the prompt, allowing GPT-4o to score the different methods ($1-5$ points). 
To mitigate the impact of the varying positions in the prompt on the scores of each method, we randomize the positions of dialogues for each method. 
We focus on evaluating whether the model can accurately diagnose the patient's disease.

\noindent \textbf{$\bullet$ Human Doctor Evaluation Metric:} 
Considering that GPT-4o may lack expertise in the medical domain, we invite human doctors (including practitioners of modern medicine and TCM) to compare different methods. 
In each evaluation case, we provide doctors with the patient's medical information and the answer generated by two different methods (e.g. our method and baseline method). 
Their task is to select the model that generates the better response.

\subsection{Experimental Results}

\begin{figure}
    \centering
    \includegraphics[width=1\linewidth]{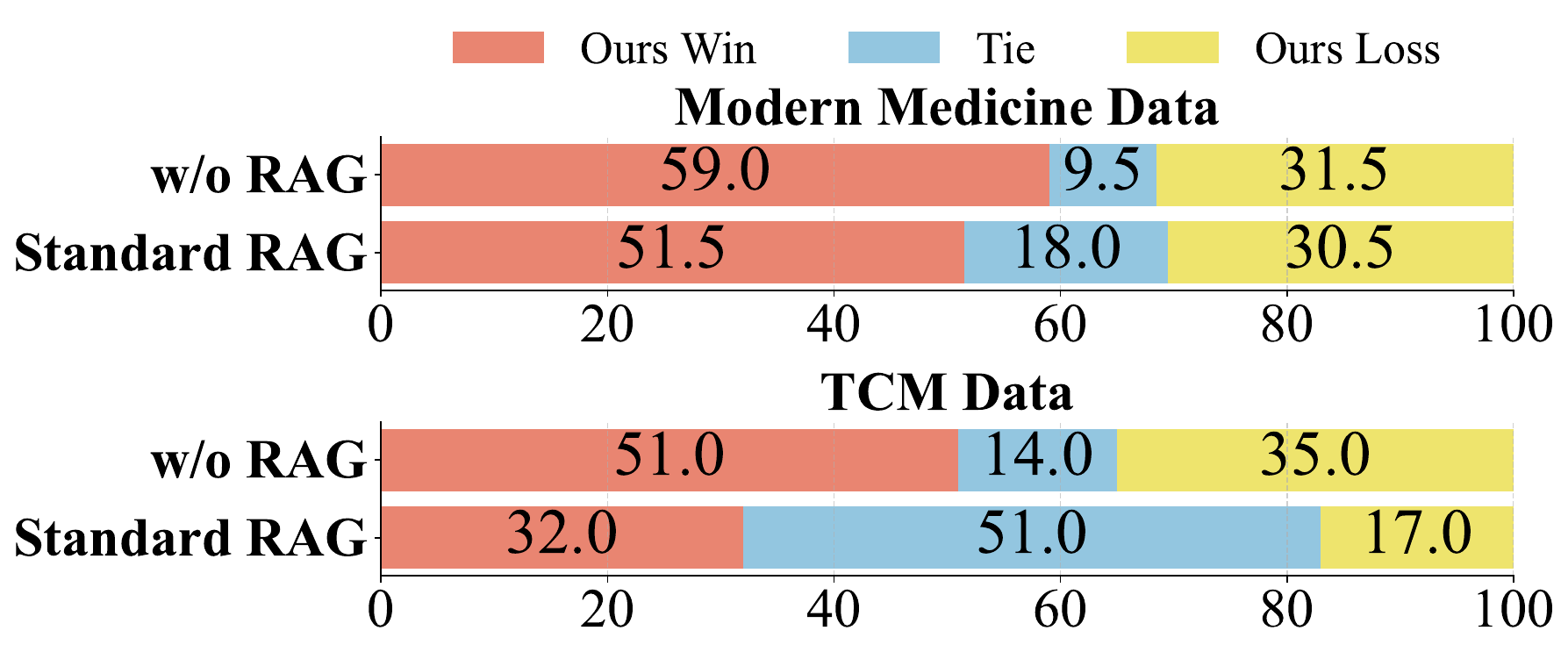}
    \caption{Comparison and evaluation results of different methods by human doctors.}
    \label{fig:Human_Eval_Results}
\end{figure}

\noindent$\bullet$~\textbf{Analysis of GPT Evaluation Metric Result.}
The results of GPT's evaluation for each method are shown in Table \ref{tab:gpt4 scores}. 
Compared to w/o RAG baselines (general LLMs and medical LLMs), all models enhanced with Standard RAG using DiagnosGraph demonstrate performance  $0.15$ gains averagely. 
This validates the effectiveness of DiagnosGraph in providing critical and accurate external evidence for the medical diagnosis task.
In comparisons with standard and medical RAG, our proposed framework  outperforms them across all evaluation metrics. 
This highlights the advantages of our architecture, particularly the dual-index retrieval mechanism and the Analyzer-Doctor architecture, which more effectively bridge the semantic gap between patient narratives and professional terminology to achieve more precise and efficient diagnostic reasoning.

\noindent$\bullet$~\textbf{Analysis of Human Doctor Evaluation Metric Result.}
We randomly select $100$ dialogues from modern medical data (CMB-Clin and MM-Cases) and $100$ from TCM data (TCM-SD-100 and TCM-Cases) for evaluation by two modern medical and two TCM doctors. 
We choose Qwen2-7B-Instruct as the base model. 
As presented in Figure~\ref{fig:Human_Eval_Results}, doctors tend to believe that our method provide more accurate diagnosis compared to the other methods. 
The average win ratio of our method compared to the LLM without RAG method is $55.00$ vs. $33.25$ ($\uparrow21.75\%$), and compared to the standrad RAG method, the average win ratio is $41.75$ vs. $23.75$ ($\uparrow18\%$). 
The evaluation results from human doctors are consistent with the previous GPT evaluation.

\subsection{Analysis of Retrieval Performance}
\label{app:Analysis_of_Retrieval_Performance}

To verify the effectiveness of $r_{MR}$, we use the first utterance of the patient in the dialogue as a query to retrieve the DiagnosGraph and compare it with $r_{DI}$. 
We use mean reciprocal rank (MRR) and Hits@n as an evaluation metric for retrieval performance.
As shown in Table \ref{tab:retrieval performance}, retrieval performance of $r_{MR}$ is improved compared to $r_{DI}$ over $0.171$ and $0.158$ in MRR and Hits@1 metric respectively. 
It suggests that $r_{MR}$ is closer to the semantics of the patient.

\begin{table}[t]
  \centering
  \caption{Retrieval performance of $r_{MR}$ and $r_{DI}$.}
    \scalebox{1.05}{
\begin{tabular}{@{}cccccc@{}}
\toprule
                            & Index Type   & MRR   & Hits@1 & Hits@3 & Hits@10 \\ \midrule
\multirow{2}{*}{CMB-Clin}   & $r_{DI}$ & 0.084 & 0.027  & 0.068  & 0.203   \\
                            & $r_{MR}$ & \textbf{0.254} & \textbf{0.135}  & \textbf{0.338}  & \textbf{0.432}   \\ \midrule
\multirow{2}{*}{MM-Cases}   & $r_{DI}$ & 0.169 & 0.094  & 0.177  & 0.312   \\
                            & $r_{MR}$ & \textbf{0.440} & \textbf{0.322}  & \textbf{0.481}  & \textbf{0.686}   \\ \midrule
\multirow{2}{*}{TCM-SD-100} & $r_{DI}$ & 0.090 & 0.020  & 0.060  & 0.240   \\
                            & $r_{MR}$ & \textbf{0.111} & \textbf{0.040}  & \textbf{0.120}  & \textbf{0.250}   \\ \midrule
\multirow{2}{*}{TCM-Cases}  & $r_{DI}$ & 0.472 & 0.308  & 0.569  & 0.792   \\
                            & $r_{MR}$ & \textbf{0.694} & \textbf{0.585}  & \textbf{0.754}  & \textbf{0.908}   \\ \bottomrule
\end{tabular}
}%
  \label{tab:retrieval performance}%
\end{table}%

\subsection{Case Study}
\label{app:case_study}

\begin{figure*}[htbp!]
    \makebox[\textwidth][c]{\includegraphics[width=0.9\linewidth]{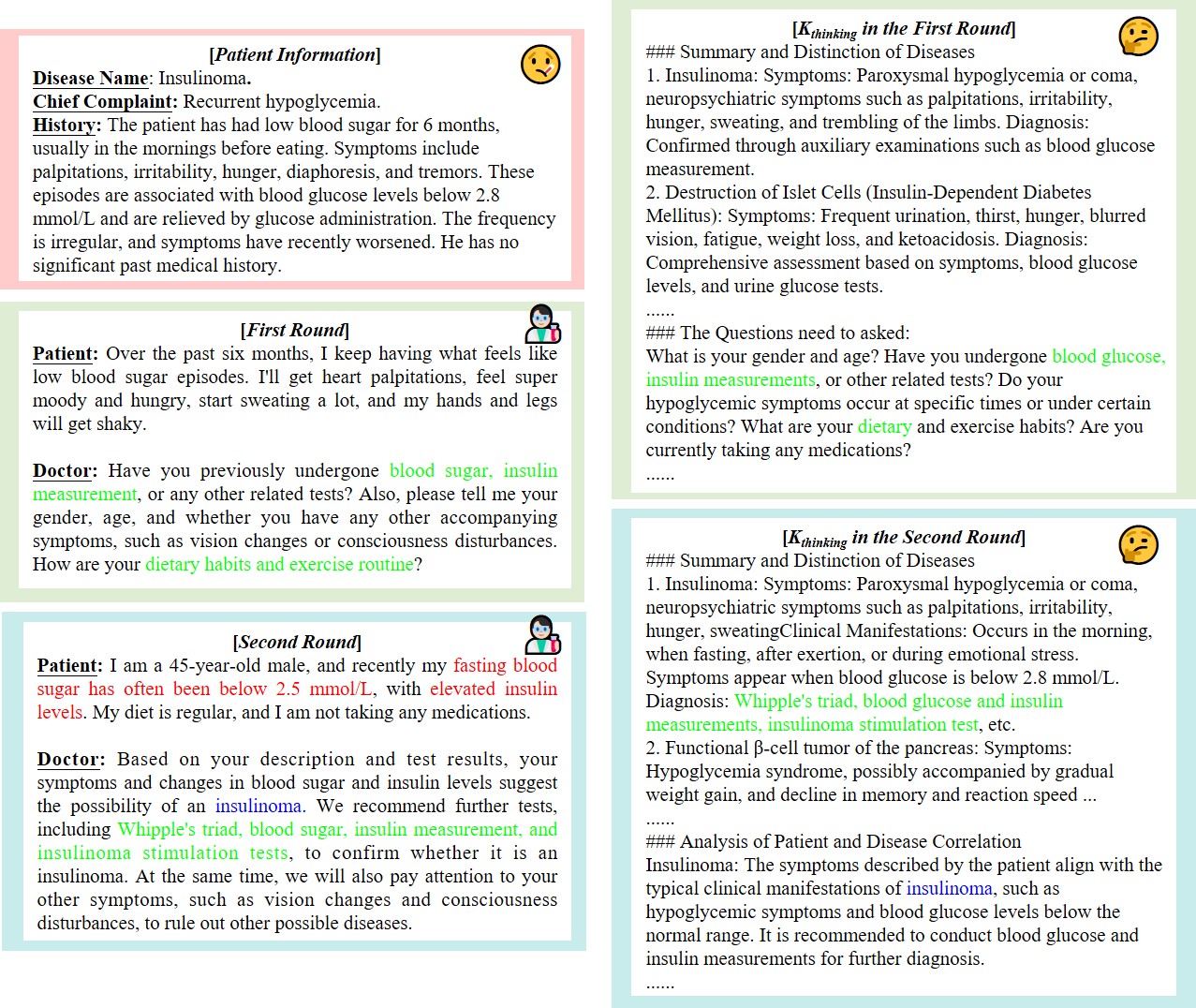}}
    \caption{A diagnostic case of MRD-RAG on the modern medicine case. \textcolor[RGB]{255,0,0}{Red} indicates keywords from the patient's utterance that contribute to the diagnosis. \textcolor[RGB]{0,255,0}{Green} indicates keywords from retrieved knowledge. \textcolor[RGB]{0,0,255}{Blue} indicates diagnosis results.}
    \label{fig:case study mm}
\end{figure*}

\begin{figure*}[htbp!]
    \makebox[\textwidth][c]{\includegraphics[width=1.0\linewidth]{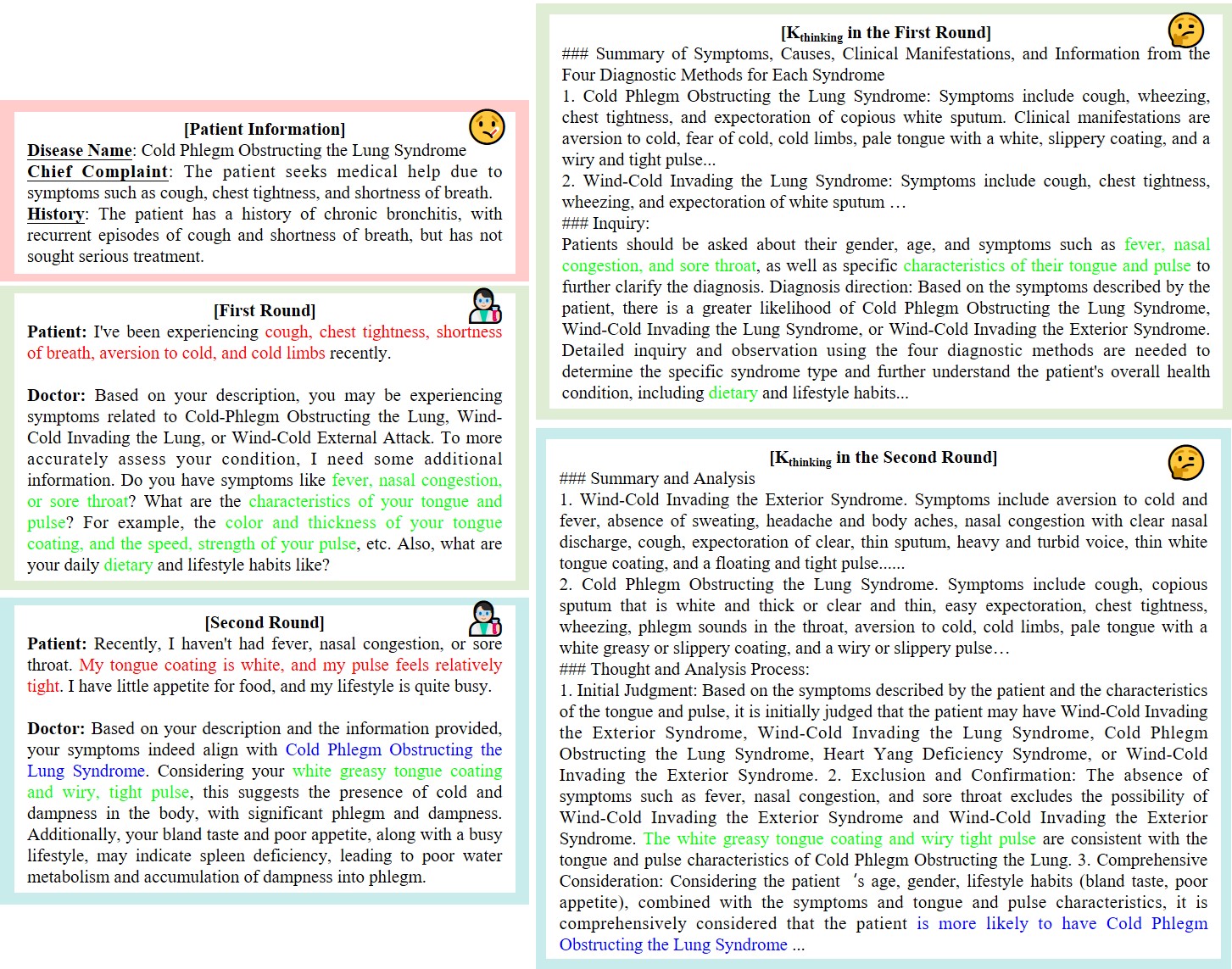}}
    \caption{A diagnostic case of MRD-RAG on the TCM case. \textcolor[RGB]{255,0,0}{Red} indicates keywords from the patient's utterance that contribute to the diagnosis. \textcolor[RGB]{0,255,0}{Green} indicates keywords from retrieved knowledge. \textcolor[RGB]{0,0,255}{Blue} indicates diagnosis results.}
    \label{fig:case study tcm}
\end{figure*}

\noindent $\bullet$ \textbf{Modern Medicine Case.} As shown in Figure \ref{fig:case study mm}, in the first round, the dialogue begins with the patient's colloquial description of "\textit{what feels like low blood sugar episodes}". 
Instead of making a premature diagnosis, the system's Analyzer module processes retrieved candidate diseases (like insulinoma) and formulates targeted questions for key clinical data, such as blood glucose and insulin measurements, which the patient may not know are important.
After the patient provides these specific lab values in the second round, the framework's reasoning process intensifies. 
The Analyzer module correlates the new, professional-level data (low fasting blood sugar with high insulin) with its knowledge base, inferring a high probability of insulinoma and identifying Whipple's triad as the next diagnostic step. 
The Doctor module then communicates this, recommending further tests to confirm the diagnosis.

\noindent $\bullet$ \textbf{TCM Case.} As shown in Figure \ref{fig:case study tcm}, the dialogue starts with the patient's general symptoms like "\textit{cough and chest tightness.}" 
Our framework retrieves several similar syndromes and, instead of guessing, the Analyzer module identifies the key differentiating factors in TCM theory. 
It then prompts the Doctor module to ask for specific, professional diagnostic information (e.g., tongue characteristics) that a layperson would not typically volunteer, effectively bridging the initial knowledge gap.
Once the patient provides this key information (the color of their tongue coating) in the second round, the Analyzer module can accurately differentiate between the candidate syndromes. 
It synthesizes the patient's narrative with the retrieved professional knowledge to pinpoint the correct diagnosis.

\subsection{Ablation Study}

\begin{figure}
    \centering
    \includegraphics[width=0.9\linewidth]{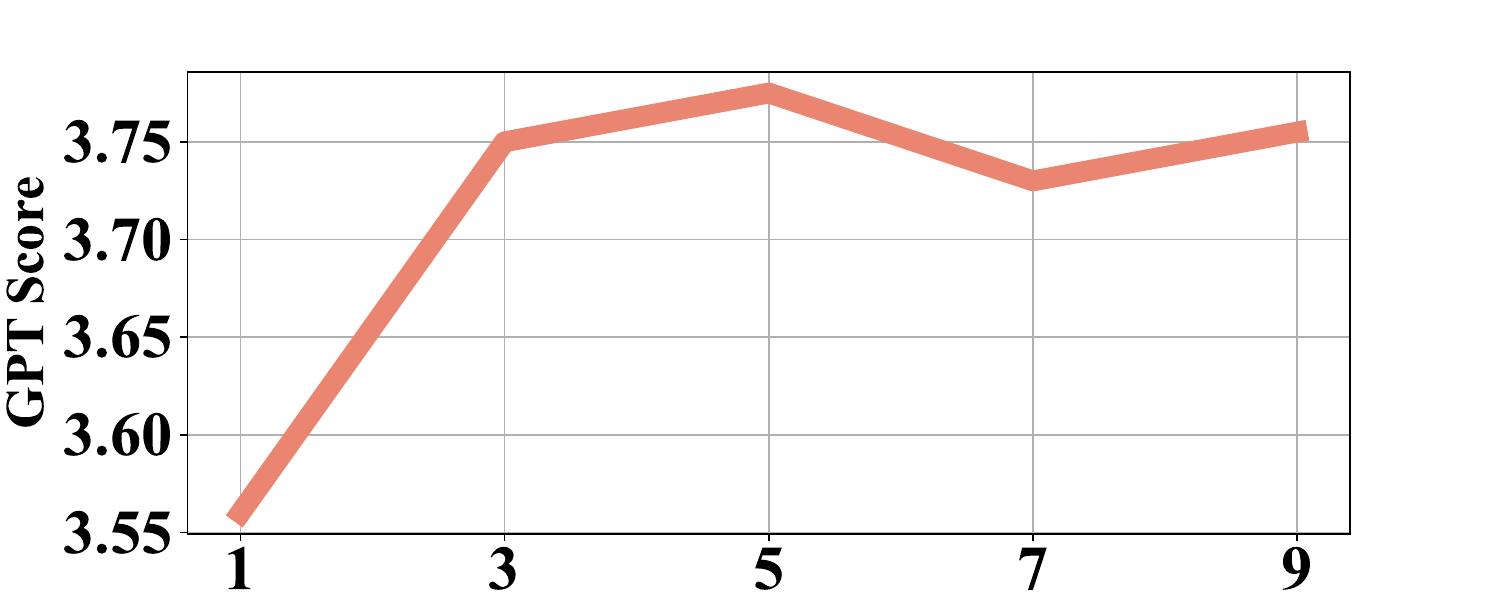}
    \caption{Ablation study with $top\text{-}k$, from 1 to 9.}
    \label{fig:topk}
\end{figure}

\noindent$\bullet$ \textbf{Top-k.} We test the changes in MRD-RAG's GPT score with different $top\text{-}k$ values. 
As shown in Figure \ref{fig:topk}, when $top\text{-}k=5$, MRD-RAG achieves the best diagnostic performance. If $top\text{-}k$ continues to increase, MRD-RAG's diagnostic performance gets worse. 
Larger $top\text{-}k$ values result in retrieving too many irrelevant diseases, which affects MRD-RAG's ability to analyze differences between diseases.

\noindent $\bullet$ \textbf{Analyzer Module.}
We performe an ablation study to validate the necessity of the Analyzer module, as shown in Table~\ref{tab:ablation_study}. 
The experiment compare the performance of MRD-RAG against a variant without the Analyzer, evaluated via a GPT scoring.
The results show that removing the Analyzer leads to a decline in model performance. 
This confirms our design hypothesis: without a dedicated module to perform differential diagnosis, the Doctor module struggles to handle complex clinical reasoning. 

\begin{table}[t]
  \centering
  \caption{Ablation study of Analyzer Module.}
    \scalebox{0.9}{
\begin{tabular}{@{}c|cccc@{}}
\toprule
Method       & CMB-Clin & MM-Cases & TCM-SD-100 & TCM-Cases \\ \midrule
w/o Analyzer & 3.02     & 2.93     & 3.26       & 3.92      \\
MRD-RAG-DI   & \textbf{3.12}     & \textbf{2.96}     & \textbf{3.34}       & \textbf{3.94}      \\ \midrule
w/o Analyzer & 3.64     & 3.7      & 3.18       & 4.38      \\
MRD-RAG-MR  & \textbf{3.74}     & \textbf{3.88}     & \textbf{3.42}       & \textbf{4.46}      \\ \bottomrule
\end{tabular}
}%
  \label{tab:ablation_study}%
\end{table}%

\section{Conclusion}
This work introduces a novel approach to mitigate the semantic gap in medical RAG from both data and methodological perspectives. 
First, we construct DignoseGraph, a knowledge base designed to map subjective patient complaints to structured medical knowledge.
Second, we propose MRD-RAG, a diagnostic framework that simulates a physician's reasoning process. 
Its architecture allows it to dynamically refine hypotheses through cycles of differential diagnosis and precise, targeted questioning.
Experimental results on four datasets, along with detailed ablation and case study analyses, validate the superiority of MRD-RAG. 
Our work demonstrates that by co-designing the knowledge resource and the reasoning framework, we can mitigate the gap between patient language and clinical terminology.


\end{document}